%% file: camera-ready.tex
\documentclass[sigconf]{acmart}

\AtBeginDocument{%
  \providecommand\BibTeX{{%
    \normalfont B\kern-0.5em{\scshape i\kern-0.25em b}\kern-0.8em\TeX}}}

\DeclareMathOperator*{\argmax}{arg\,max}
\DeclareMathOperator*{\argmin}{arg\,min}
\usepackage{multirow} 
\usepackage{natbib}  
\usepackage{subcaption}
\usepackage{header} 
\usepackage[draft]{fixme}
\fxsetup{inline,nomargin,theme=color}
\DeclareMathAlphabet{\mathcal}{OMS}{cmsy}{m}{n}

\copyrightyear{2020}
\acmYear{2020}
\setcopyright{rightsretained}
\acmConference[KDD '20]{Proceedings of the 26th ACM SIGKDD Conference on Knowledge Discovery and Data Mining}{August 23--27, 2020}{Virtual Event, CA, USA}
\acmBooktitle{Proceedings of the 26th ACM SIGKDD Conference on Knowledge Discovery and Data Mining (KDD '20), August 23--27, 2020, Virtual Event, CA, USA}
\acmDOI{10.1145/3394486.3403245}
\acmISBN{978-1-4503-7998-4/20/08}

\begin{document}
\fancyhead{}

\title{Treatment Policy Learning in Multiobjective Settings with Fully Observed Outcomes}

\author{Soorajnath Boominathan}
\email{soorajb@mit.edu}
\affiliation{
  \institution{MIT CSAIL, IMES}
  \city{Cambridge}
  \state{MA}
  \postcode{02139}
}
\author{Michael Oberst}
\email{moberst@mit.edu}
\affiliation{
  \institution{MIT CSAIL, IMES}
  \city{Cambridge}
  \state{MA}
  \postcode{02139}
}

\author{Helen Zhou}
\email{hlzhou@andrew.cmu.edu}
\affiliation{
  \institution{Carnegie Mellon University}
  \city{Pittsburgh}
  \state{PA}
  \postcode{02139}
}
\author{Sanjat Kanjilal}
\email{skanjilal@bwh.harvard.edu}
\affiliation{
  \institution{Harvard Medical School}
  \institution{Harvard Pilgrim Healthcare Institute}
  \city{Boston}
  \state{MA}
  \postcode{02139}
}
\author{David Sontag}
\email{dsontag@csail.mit.edu}
\affiliation{
  \institution{MIT CSAIL, IMES}
  \city{Cambridge}
  \state{MA}
  \postcode{02139}
}

\begin{abstract}

In several medical decision-making problems, such as antibiotic prescription, laboratory testing can provide precise indications for how a patient will respond to different treatment options. This enables us to ``fully observe'' all potential treatment outcomes, but while present in historical data, these results are infeasible to produce in real-time at the point of the \emph{initial} treatment decision.  Moreover, treatment policies in these settings often need to trade off between multiple competing objectives, such as effectiveness of treatment and harmful side effects. 

We present, compare, and evaluate three approaches for learning individualized treatment policies in this setting:  First, we consider two \emph{indirect} approaches, which use predictive models of treatment response to construct policies optimal for different trade-offs between objectives.  Second, we consider a \emph{direct} approach that constructs such a set of policies without intermediate models of outcomes.  Using a medical dataset of Urinary Tract Infection (UTI) patients, we show that all approaches learn policies that achieve strictly better performance on all outcomes than clinicians, while also trading off between different objectives.  We demonstrate additional benefits of the direct approach, including flexibly incorporating other goals such as deferral to physicians on simple cases.

\end{abstract}

\ccsdesc[500]{Computing methodologies~Supervised learning}
\ccsdesc[300]{Applied computing~Health care information systems}

\keywords{Healthcare, machine learning, policy learning, decision making, antibiotics, learning to defer, multi-objective optimization}

\maketitle

\section{Introduction}

Many medical treatment settings involve time-intensive or expensive laboratory testing that provide information about patient responses to all treatments of interest, but which are unavailable to doctors when they must first make a treatment decision. For instance, in antibiotic prescription, patients are tested for resistance to several antibiotics, not just those used for treatment, providing us with information about their response to all relevant treatments. However, these results take days to obtain, and doctors have to make an immediate treatment decision in the meantime. 

A similar setting arises in cancer treatment. In patient-derived xenografts (PDX) models, a patient's tumor tissue is implanted in a mouse, where its response to various cancer treatments can be recorded. However, these models take 2-8 months to develop, so physicians typically need to start patients on treatment without knowing the optimal choice \citep{pmid31725451}. 

Treatment decisions in many medical settings also require doctors to make trade-offs between competing objectives. For instance, a doctor might aim to maximize treatment effectiveness while constraining the risk of adverse side-effects to the patient or overall treatment cost. In the case of antibiotic prescription, doctors have to make a trade-off between the objectives of treatment effectiveness and minimizing usage of 2nd-line, or broad spectrum, antibiotics. Broad spectrum antibiotics have a higher likelihood of being effective, but overuse leads to increased population-level resistance rates in the long run. In this setting, the objective is to maximize antibiotic effectiveness while limiting usage of 2nd-line antibiotics. 

In this work, we present methods for learning treatment policies in such multi-objective settings with fully observed outcomes for the treatments of interest. 
We present both indirect and direct policy learning approaches for learning a set of treatment policies that exhibit various trade-offs between the objectives. We then apply these methods to learn policies for making antibiotic prescription decisions in patients with urinary tract infections (UTIs). Our primary contributions in this work are as follows:
\begin{itemize}
    \item We present three treatment policy learning methods for settings with fully observed outcomes and multiple objectives, and highlight the trade-offs across the various approaches.
    \item Using a medical dataset of Urinary Tract Infection (UTI) patients, we show that all methods are able to significantly exceed clinician performance with respect to multiple important treatment objectives.
    \item We show that our methods are able to learn a set of policies that can effectively trade off between treatment objectives, enabling practitioners to select a policy with the desired trade-off at decision time.
    \item We use a synthetic environment to demonstrate that direct policy learning learns superior treatment policies relative to indirect approaches in settings with complex outcome models, but simple optimal treatment rules.
    \item We show that the direct learning framework naturally accommodates other considerations, such as deferral to clinician decisions in situations where physicians typically perform well.  We demonstrate this use case in the antibiotic prescription setting as well.
\end{itemize}
Code for the methods presented in this paper can be found at \url{https://github.com/clinicalml/fully-observed-policy-learning}.
\section{Related Work}

There are several related areas of research that deal with learning policies from retrospective data of contexts, actions, and outcomes, and we outline some connections here.  Several lines of research focus on the setting where the only observed outcome corresponds to the action that was taken.  In the contextual bandits literature, this is referred to as bandit feedback, and the retrospective setting is referred to as the ``batch'' setting \citep{Swaminathan15a}.  In biostatistics, epidemiology, and causal inference more broadly, the outcomes are referred to as ``potential'' or ``counterfactual'' outcomes \citep{Imbens2015, Hernan2020} and the fact that we only observe a single outcome per individual is referred to as the fundamental problem of causal inference.  

Within epidemiology and biostatistics, a policy is sometimes referred to as an individualized treatment rule (ITR), and two broad approaches exist to learning them, an \textit{indirect} and a \textit{direct} approach.  In the indirect approach, the conditional distribution of the outcome (given patient characteristics) for each action is modeled directly, and a decision rule is obtained by choosing the action that maximizes the outcome of interest.  This includes approaches such as Q-learning, A-learning, and regret regression  \citep{q-learning-orig, qa-learning, regret-regression}.
However, the model class of the resulting ITR is dependent on the model class used for the conditional outcomes:  If a linear ITR is desired, then it necessitates the use of a linear model class for the conditional outcome models.  In cases where a simple outcome model does not accurately capture the true conditional outcome function, the learned ITR may be sub-optimal \citep{itr-performance}. By contrast, the \textit{direct} approach develops an estimator of the expected conditional outcome as a function of a decision rule, and directly optimizes the value of this estimator by searching over the space of treatment decision rules.  This removes the dependence between the complexity of the learned ITR and the outcome models to avoid misspecification.  For problems with binary treatments, this direct optimization problem is equivalent to weighted binary classification, and is referred to as outcome weighted learning \citep{owl}. Recent work has also focused on developing and analyzing convex surrogates for this problem to allow for efficient optimization of the objective \citep{earl} and extend this to the multi-action setting \citep{Huang2019}.  

In our setting, we have access to counterfactual outcomes of all possible treatments.  Thus, a close setting to ours is that of cost-sensitive multi-class classification \citep{Elkan2001}, where the true loss function is the multi-class equivalent of a weighted 0-1 loss in binary classification.  In this setting, smooth convex surrogates for this objective are used for learning \citep{Zou2008}, with the desideratum that they are consistent for the Bayes optimal classifier \citep{Zhang2004, Tewari2007}, and \citet{Huang2019} use the same multinomial deviance risk that we use as our direct policy learning approach.  All of these cases, as well as our own, are concerned with deterministic policies, because these are generally optimal for cases where exploration is not required (and stochastic policies introduce further optimization difficulties, as studied in the bandits literature \citep{Chen2019}).

Multi-objective decision-making has been long-studied in the context of single decisions \citep{Zeleny1982, Vira1983} and recently in the context of sequential decision-making, typically formulated as a Markov Decision Process with a vector-valued objective \citep{Roijers2013} and a \textit{scalarization function} (to convert the vector-valued objective into a scalar reward) that is unknown at train-time, but often assumed to be a linear combination \citep{Natarajan2005}.  Many of these methods seek to maintain a set of policies that are optimal for different linear scalarization functions \citep{Natarajan2005,Barrett2008,Lizotte2012}, and can be seen as indirect methods in their use of Q-functions to do so.  The set of objective values achieved by the set of optimal policies (each optimal for a different possible scalarization function) is known as the \textit{Pareto frontier} \citep{Yang2019}.  Our indirect approach of expected reward maximization can be seen as roughly the one-step (and therefore much simpler) equivalent of some of these methods.  Our direct approach borrows (albeit more conceptually) from these methods as well, in that we learn a set of policies corresponding to a set of fixed linear preferences.

In our specific application area of antibiotic prescription, recent work \citep{Yelin2019} learned treatment policies in a setting with fully observed outcomes, but with the single objective of maximizing treatment effectiveness. They used an indirect approach, training models to predict resistance and selecting the antibiotic with the minimum predicted resistance probability. Their work did not address the multi-objective nature of the antibiotic prescription problem. Another work \citep{abx-thresholding} developed models for predicting resistance to a single antibiotic using a utility-based objective that accounted for factors beyond treatment effectiveness, such as drug cost. However, they do not address how these resistance predictions or their proposed utility-based framework could be used to construct a treatment policy that selects among several antibiotics.

\section{Policy Learning Methods}
\label{sec:methods}

\subsection{Overview}

In the general formulation of the multi-objective policy learning problem, we let $\cA = [K]$ denote the action space, where $K$ is the number of discrete actions, we denote features as $\bX \in \R^m$, and we will seek to choose a policy $\pi: \R^m \rightarrow \cA$ which maps from features (i.e., patient characteristics) to recommended actions. Note that we use bold-faced symbols like $\bX$ to denote vectors, and $X(i)$ to denote the $i$-th entry of a vector. 

In our setting, we focus on optimizing over two objectives, and adapt our notation accordingly.  Our dataset is of the form $\lbrace (\textbf{X}_i, \textbf{Y}_i, \textbf{C}_i) \rbrace_{i=1}^n$, where $\textbf{X}_i \in \mathbb{R}^m$ are patient features, and $\bY_i, \bC_i$ represent our competing objectives, a \textit{benefit} and a \textit{cost} respectively.  In the antibiotic prescription setting, $\textbf{Y}_i \in \lbrace 0, 1 \rbrace^K$, where $Y(a)$ is an indicator for whether antibiotic $a$ is effective in treating an infection, and $\textbf{C}_i \in \{0, 1\}^K$, where $C(a)$ is an indicator for whether the chosen antibiotic is broad spectrum, whose use we wish to minimize in the interests of avoiding population-level antibiotic resistance.  We make two remarks:  First, broad spectrum antibiotics tend to be more effective in expectation, leading to a trade-off between these two objectives.  Second, we observe outcomes for \emph{all} treatments, not just the one received by a patient, as our dataset contains laboratory results that indicate the susceptibility of the infection to different drugs.

We present three approaches in this section.  Each approach will seek to return a \textbf{set} of policies $\Pi$, where each element $\pi \in \Pi$ represents an optimal policy for some trade-off between $Y$ and $C$.  The first two methods can be viewed as \emph{indirect} approaches, in that they require learning a separate model for the conditional mean of $Y$ under each treatment,\footnote{In our case, costs are determined by the choice of treatment itself, so we only need to model the conditional mean of treatment effectiveness, but it is straightforward to extend both methods to the case where all objectives must be modelled.} denoted $f_a(x) \approx \E(\textbf{Y}(a) \mid \textbf{X}=x)$.  The third approach is a \emph{direct} approach, in that it does not require predictive models for the individual outcomes, but optimizes directly for a treatment policy.  To summarize these approaches:

\begin{enumerate}
    \item \textbf{Thresholding}: Use a set of carefully-chosen thresholds to convert $f_a(x)$ into a binary prediction of effectiveness $Y(a)$ for each treatment, and then choose the lowest-cost treatment which is predicted to be effective.
    \item \textbf{Expected Reward Maximization}: Combine $Y$ and $C$ into a single objective $r_{\omega}$ (the ``reward'') by taking linear combinations with varying weights, and choose the treatment which maximizes this reward according to the models $f_a(x)$.
    \item \textbf{Direct Policy Optimization}: Using the same definition of reward $r_{\omega}$, learn a single model that directly predicts which treatment is optimal by optimizing a surrogate loss.
\end{enumerate}

\subsection{Thresholding}
\label{sub:thresholding}

In this section, we introduce a simple method whose decision logic is intuitive: Predict which treatments will be effective, and then choose the effective treatment with the lowest cost.  Given our learned models $f_a(x)$ of predicted effectiveness, which output numbers between $0$ and $1$, we use carefully-chosen thresholds to make a binary prediction for each $Y(a)$.  We combine these predictions with the fixed cost associated with broad spectrum antibiotics to choose the lowest-cost treatment that is still predicted to be effective.

More formally, we denote the set of thresholds used to binarize each prediction as $\lbrace t_a \rbrace_{a=1}^K$. We let $e_a(x)$ be an indicator\footnote{Throughout, we use $\1{P}$ as an indicator function that is equal to 1 if the expression $P$ is true, and $0$ if $P$ is false} that represents whether treatment $a$ is predicted to be effective for patient with features $x$, given by
\begin{equation}
    e_a(x) = \1{f_a(x) \geq t_a}.
\end{equation}
The treatment policy is then defined as the action that minimizes cost, among the treatments that are predicted to be effective
\begin{equation}
    \pi(x) = \argmin_{a} \lbrace C(a) \mid e_a(x) = 1\rbrace.
\end{equation}
If $e_a(x) = 0$ for all $a \in \cA$, the treatment policy falls back to an option $a$ that minimizes the cost $C(a)$.  In the setting of antibiotic prescription, this corresponds to defaulting to a 1st-line antibiotic.

\textbf{Choosing thresholds}: A single set of thresholds implicitly defines a policy with a fixed trade-off between effectiveness and other costs.  To construct a set of policies $\Pi$ that express varying preferences between treatment effectiveness and other costs, we perform an exhaustive search over different choices of threshold combinations $\lbrace t_a \rbrace_{a=1}^K \in \cT$, where $\cT$ is a large (but finite) search space.  

Each policy $\pi$ implied by our models $f_a$ and thresholds $t_a$ is then evaluated on a held-out validation set to get an empirical estimate of their expected benefit $\E[Y(\pi(x))]$ and expected cost $\E[C(\pi(x))]$.  We then enumerate over a set of cost constraints $\{ b_j \}_{j=1}^{J}$, and return the $J$ policies which satisfy
\begin{equation}
    \pi^* = \argmax_{\pi} \{ \E[Y(\pi(x))] : \E[C(\pi(x))] \leq b_j\}.
\end{equation}
In other words, we choose the policy for each $b_j$ which achieves the highest mean value of $Y$ in our validation set, subject to the constraint that the mean cost is less than $b_j$.

This approach, while straightforward and interpretable, does have drawbacks.  It requires enumeration over a large set of thresholds $\cT$, and thresholding predicted probabilities throws away information:  For instance, two treatments with equal cost could both have predicted probabilities of effectiveness greater than their respective thresholds, but where the model is far more confident in one over the other.  In the next section, we present a method that circumvents these issues, while making the trade-off that the resulting decision logic (maximizing an expected reward) may be less interpretable to a lay audience.

\subsection{Expected Reward Maximization}
\label{sub:expected_reward_max}

In single objective settings, a simple approach for converting predictions of treatment effectiveness into a policy for a patient with features $x$ is to select the treatment $a^* = \argmax_a f_a(x)$, i.e., the treatment that is most likely to work, based on model predictions.

Our setting has multiple objectives, but if we combine our objectives into a single number indicating a notion of value or reward, then we can construct a similar policy that optimizes this quantity.

More formally, recall that our goal is to learn a deterministic treatment policy $\pi: \mathbb{R}^m \rightarrow \cA$, which maps patient features to a deterministic decision. To combine our objectives, we will use a linear preference parameter $\omega \in [0, 1]$ such that the reward is linear combination of our competing objectives.  In our particular case, we parameterize this as follows, to account for the fact the our cost is a binary variable
\begin{equation}
    \textbf{r}_{\omega} = \omega \cdot \textbf{Y} + (1 - \omega) \cdot (1 - \textbf{C}),
    \label{eq:reward_omega}
\end{equation}
where $\textbf{r}_{\omega}(a)$ represents the reward under treatment $a$.  We will omit the subscript where it is clear from context.

In the setting of linear preferences, commonly used in the multi-objective optimization literature \citep{Stewart1992, Natarajan2005} we do not lose anything by restricting ourselves to deterministic policies, because there exists an optimal policy that is deterministic \citep{Roijers2013}. Under this preference $\omega$, we define the Bayes optimal policy $\pi_{\omega}^*$ as the one that maximizes the expected reward for a given $x$, and is defined with respect to the true (unknown) conditional expectations
\begin{equation}
    \pi_{\omega}^*(x) = \argmax_{a \in A} \mathbb{E}[r_{\omega}(a) \mid x].
\end{equation}
In the setting we consider, we adopt a strategy of learning a \emph{set} of policies $\Pi$ that are each optimal according to some preference $\omega$, allowing users to select a policy from this set which corresponds to their desired trade-off.  This is referred to as the ``decision support'' setting \citep{Roijers2013} in that we do not solicit an explicit preference $\omega$ from decision makers, but instead provide a set of alternatives that are each optimal for some $\omega$.

Using this formalism and the definition of reward in Equation~\eqref{eq:reward_omega}, we can use our models $f_a(x)$, which approximate $\E[Y(a) | X = x]$, to construct a prediction of this reward under each action $a$, and then define our treatment policy $\pi_{\omega}(x)$ as the one that chooses the action with the highest predicted reward
\begin{equation}
    \pi_{\omega}(x) = \argmax_a \omega \cdot f_a(x) + (1 - \omega) \cdot (1 - C(a)). 
\end{equation}%
Constructing such a decision rule for each $\omega$ produces our desired set of policies $\Pi$.  This approach has the benefit of not requiring enumeration over a large set of thresholds, and it takes the predicted probabilities into account directly.  

However, this approach requires us to build predictive models of treatment effectiveness, and can introduce a trade-off between policy performance and interpretability. For instance, representing the outcome models $f_a$ with linear functions allows us to interpret the learned policy and gain insight into features driving decisions by examining differences in coefficients.\footnote{If there are two actions, then this is a direct consequence of the formulation, and for more than two actions the policy can be interpreted as a set of linear classifiers by comparing the difference in coefficients for models of pairs of treatments.}  In many settings, linear models may be too simple to accurately model outcomes, which can lead to models with poor predictive performance (and therefore sub-optimal policies). On the other hand, more complex models sacrifice the interpretability of the resulting policy. In the next section, we present a method which instead seeks to find a policy of the desired model class (e.g., linear) directly.

\subsection{Direct Policy Optimization}

In this approach, we seek to directly learn a policy which has an interpretable form, without learning any specific models of treatment effectiveness.  We use the same notion of reward defined in Section~\ref{sub:expected_reward_max}, and optimize the (estimated) value of a treatment policy $\pi$, $\hat{V}_{\omega}(\pi)$.  As before, we learn a range of policies corresponding to different values of $\omega$. The value of a policy $V_{\omega}(\pi)$  is defined as
\begin{equation}
    V_{\omega}(\pi) 
  \coloneqq \E_{x,r} \left[ \sum^{}_{a \in \cA} r_{\omega}(a) \1{a = \pi(x)}\right],
\end{equation}
and $\hat{V}_{\omega}(\pi)$ is the empirical estimate of this quantity. In this case, our goal is to find a function $\pi: \R^m \rightarrow \cA$ which maximizes this objective. We note that any such policy can be written as $\pi(x) = \argmax_{a \in \cA} d(x, a)$ for some function $d: \R^m \times \cA \rightarrow \R$. The optimal policy $\pi_{\omega}^*$ can then be written as
\begin{equation}
  \pi_{\omega}^* = \argmax_{\pi} \E_{x,r} \left[ \sum^{}_{a \in \cA} r_{\omega}(a) \1{a = \argmax_{a \in \cA} d(x, a)} \right].
\end{equation}
We  omit $\omega$ in the remainder of this section, as we will choose a finite set of values for $\omega$ to generate a set of optimal policies. 

To find an optimal policy, we wish to optimize over the space of decision functions $d$ using the empirical estimate of $V(\pi)$, but the argmax operation causes this objective to be non-convex.  Instead, we use a differentiable \textit{convex surrogate} objective \citep{Tewari2007, Zou2008}, in our case the multinomial deviance loss \citep{Zou2008, Huang2019}, which has the appealing property that it is not only convex and differentiable, but that when solved to optimality it yields a consistent estimator of the Bayes-optimal policy.  Concretely, we optimize over functions $f_a: \R^m \rightarrow \R$, where our resulting policy will be given by $\pi(x) = \argmax_{a} f_a(x)$, and minimize the following quantity in our empirical sample
\begin{equation}
  \E_{x, r} \tilde{L}(f, x, r) \coloneqq - \E \left[ \sum^{}_{a \in \cA} r(a) \log \frac{\exp f_a(x)}{\sum_{a'} \exp f_{a'}(x)} \right].
  \label{eq:convex_surrogate}
\end{equation}
In this work, we parameterize the functions $f_a$ with a linear model, such that $f_a(x) = \theta_a^T x$. As noted, this objective has the appealing property that, when solved to optimality, it will yield a policy that is consistent for the Bayes-optimal policy in the following sense (proof provided in the appendix).
\begin{thmprop}
  For nonnegative rewards $\mathbf{r}$, and for an $f^*$ that satisfies $f^* = \inf_f \E_{x, r} \tL(f, x, r)$, the corresponding policy $\pi^*(x) = \argmax_a f^*_a(x)$ is equivalent to the the Bayes-optimal policy $\pi^*(x) = \argmax_a \E[r(a) | X]$.
  \label{prop:consistent_optimal}
\end{thmprop}

Directly optimizing for a treatment policy in this way decouples the complexity of the outcome models in a given setting from the complexity of an effective treatment policy. This enables the learning of interpretable decision-making policies even in a setting where modeling outcomes requires extremely sophisticated models.

\section{Experiments}
When a patient presents with an infection, clinicians typically need to make an immediate decision on which antibiotic to administer.  Laboratory tests can assess the susceptibility of an infection to each antibiotic, but they take time:  The bacteria must be grown in culture to sufficient quantities where the effectiveness of antibiotics can be tested directly, and this process takes several days to return results.

This is known as the `empiric treatment setting', where doctors must make the initial prescription decision without the benefit of tests, using the patient's medical history and their own clinical experience. They must weigh two competing concerns: Avoiding an inappropriate antibiotic therapy (IAT) where the pathogen is resistant to the antibiotic, while also avoiding overuse of broad spectrum therapies that lead to higher resistance rates at an individual and population level and have a higher risk of adverse side effects \cite{lee2015aortic,dingle2016cdificile}.

This problem is particularly prevalent in urinary tract infections (UTIs), a common class of infection with more than 150 million annual cases worldwide \citep{stamm2001urinary}. Resistance rates to commonly prescribed agents in UTIs can exceed $20\%$, highlighting the difficulty of prescribing an effective antibiotic \citep{uk-study}. Doctors frequently use 2nd-line antibiotics \citep{kabbani} because they are more effective at the individual level, but this contributes to higher prevalence of antibiotic-resistant pathogens in the future at the population level.  Conversely, 1st-line (narrow spectrum) antibiotics are more likely to be ineffective at an individual level, but pose less risk of increasing population-level resistance. 

In practice, this trade-off is difficult to balance. Antibiotic resistance is influenced by a wide range of risk factors, from individual history of infection to population-level resistance rates and antibiotic usage trends, which are difficult to incorporate in real time decision-making. This motivates the need for learning antibiotic treatment policies that can make this trade-off in an optimal fashion.

\subsection{Data}

Our dataset is derived from the electronic health record (EHR) of Massachusetts General Hospital and the Brigham \& Women's Hospital in Boston, MA, containing the full medical record for every patient who has undergone an antibiotic resistance test.  This study was approved by the Institutional Review Board of Massachusetts General Hospital with a waived requirement for informed consent. 

\subsubsection{Cohort} Our cohort for this work consists of 15,806 microbiological specimens collected from 13,682 women with UTIs between 2007 and 2016. We filter for patients with \textit{uncomplicated} UTI, which refers to an infection in women who are not pregnant, have a structurally normal urinary tract, and have not had surgical procedures in the last 90 days. Patients with uncomplicated UTI typically receive prescriptions from a well-defined set of antibiotics, allowing us to clearly specify our action space for policy learning and evaluate performance of learned policies against clinician decisions. 

\subsubsection{Available Data} The dataset contains demographics (e.g., age, race), medications (including antibiotic prescriptions), basic lab test results, medical procedures, and comorbidities for each patient. It also contains information about the date and location that a specimen was collected, ground-truth antibiotic resistance profiles, and previous resistance test results or infections in a patient's history.

\subsubsection{Treatments} Our treatment space consists of common antibiotics used in treating UTIs:  nitrofurantoin (NIT), trimethoprim-sulfamethoxazole (SXT), ciprofloxacin (CIP), and levofloxacin (LVX). NIT and SXT are 1st-line (narrow spectrum) antibiotics, while CIP and LVX are 2nd-line (broad spectrum) antibiotics. We filter for uncomplicated UTI specimens treated with exactly one of these four agents in the empiric treatment setting, defined as the period spanning 2 days before to 1 day after the date of specimen collection. Information about clinician prescriptions is not used during policy learning, as our goal is not to imitate clinician actions, but to improve on them. We also filter out specimens missing resistance test results for any of these four antibiotics, since this information is necessary for a full evaluation of policy decisions.

\subsubsection{Feature Construction} We directly use age and race as features in the model. We construct binary features for antibiotic exposures, prior antibiotic resistance, prior infections, comorbidities, and procedures over the 7, 14, 30, 90, and 180-day periods preceding the specimen collection date, as well as features for \textit{any} history of prior antibiotic resistance or exposures, regardless of time. The results of common lab tests are averaged over the same time windows. 

We also construct a population-level feature called \textit{colonization pressure}, defined as the proportion of resistant specimens to a given antibiotic within a specified location and time window. Finally, we compute cumulative antibiotic usage rates across both hospitals in our data within the 90-day window preceding specimen collection, normalized by total patient volume. 

\subsubsection{Train / Test Split} Out of the total 15,806 specimens, our training set consists of 11,865 specimens collected from 2007-13, and we evaluate the learned policies on a held-out test set consists of 3,941 specimens from 2014-16.  This is after removing any specimens from the test set that came from patients who were also present in the training set to avoid any data leakage between train and test sets. Table \ref{tab:commands} contains cohort statistics, including resistance rates and the distribution of empiric prescriptions in train and test sets.
\begin{table}
  \caption{Cohort Statistics}
  \label{tab:commands}
  \centering
  \begin{tabular}{cccc}
    \toprule
    &  & \textbf{Train (2007-13)} & \textbf{Test (2014-16)}  \\
    \midrule
     & $n$ & 11,865 & 3,941 \\
    \midrule
    
    & Age & 34.1 (10.8) & 33.6 (11.1) \\
    & \% White & 64.0\% & 62.3\%  \\
    \midrule 
    \multirow{4}{*}{\textbf{\% Resistant}}& NIT & 11.2\% & 11.0\% \\
    & SXT & 19.6\% & 19.6\% \\
    & CIP & 5.3\% & 6.4\% \\
    & LVX & 5.1\% & 6.5\% \\
    \midrule 
    \multirow{4}{*}{\textbf{\% Prescribed}}
    & NIT & 15.9\% & 34.5\% \\
    & SXT & 41.5\% & 32.0\% \\
    & CIP & 39.2\% & 32.5\% \\
    & LVX & 3.3\% & 1.0\% \\
    \bottomrule
  \end{tabular}
\end{table}

\subsection{Experiment Setup and Evaluation}
Treatment policies are learned using a training set of specimens from 2007-13, and evaluated on a test set of specimens from 2014-16. Hyperparameters are tuned using twenty 70\% / 30\% train/validation splits of the training data in all approaches.

\subsubsection{Thresholding and Expected Reward Maximization}
We train logistic regression models to predict treatment effectiveness for each of the four antibiotic treatment options: NIT, SXT, CIP, and LVX. Regularization type and strength are tuned using the validation set. Using more complex, nonlinear model classes did not have a significant impact on predictive performance, so we chose not to use them in experiments.  For the thresholding-based approach, the search space of thresholds $\cT$ for each model was chosen for a diversity of false-positive rates (for prediction of susceptibility), using ROC curves on the training set, and $\cT$ is defined as all possible combinations of these threshold values. 

\subsubsection{Direct Policy Optimization}
The surrogate loss function~\eqref{eq:convex_surrogate} is optimized with the Adam optimizer and L2 regularization, and the number of training epochs is chosen with an early stopping criteria based on the mean reward on the validation set. 

\subsubsection{Evaluation}
We evaluate the learned policies with respect to two primary outcomes: IAT rate and 2nd-line usage. The IAT rate is the proportion of specimens treated with an antibiotic to which they are resistant, and 2nd-line usage is the proportion of specimens treated with either CIP or LVX.  We compute clinician performance and the performance of our policy using the recorded results of antibiotic resistance tests to obtain head-to-head comparisons.

We also compare our policies to the unconstrained and constrained policy learning approaches for antibiotic prescription presented in \cite{Yelin2019}. Unlike our approaches, these baselines only learn a single policy, not sets of policies that trade off across the multiple objectives. The unconstrained approach selects the treatment with the lowest predicted probability of resistance from the learned conditional outcome models, without constraining the rate of 2nd-line treatment selection. The constrained approach adjusts the predicted resistance probabilities by adding treatment-specific costs to each prediction, then selects the treatment minimizing this adjusted score. The costs are selected to constrain the empirical treatment distribution of the learned policy to match that of clinicians. 

\subsection{Reward Function}
\label{sub:exp_reward_defn}
We recall the definition of the composite reward function given in Section~\ref{sub:expected_reward_max} for the indirect expected reward maximization and direct policy learning approaches. The treatment effectiveness vectors $\textbf{Y}$ correspond to a patient's susceptibility to each antibiotic $Y_i(a) = \1{\text{patient } i \text{ is susceptible to antibiotic } a }$, the cost vector $\bC$ for the treatments are a function of the class of the chosen antibiotic $ C_i(a) = \1{a \text{ is a 2nd-line antibiotic}}$, and the composite treatment reward is a linear combination of the effectiveness and costs for each antibiotic using the preference $\omega \in [0, 1]$, given by $\textbf{r}_i = \omega \cdot \bY_i + (1-\omega) \cdot (1 - \bC_i)$. As $\omega$ is reduced, more weight is placed on avoiding 2nd-line antibiotic usage, even at the cost of additional cases of IAT. By varying $\omega$, we can learn a set of treatment policies that achieve different trade-offs between treatment effectiveness and broad spectrum usage. 

\subsection{Results}
\begin{table}
  \caption{Policy Comparison: Constant 2nd-line usage}
  \label{tab:comparison-2}
  \centering
  \begin{tabular}{lcc}
    \toprule
     & \textbf{IAT} & \textbf{2nd-line usage}  \\
    \midrule
    Doctor & 11.9\% & 33.6\%\\
    Thresholding & 8.9\% & 33.1\%   \\
    Expected reward maximization & 9.0\% &  28.2\% \\
    Direct learning & 8.9\% &  30.4\%\\
    Constrained selection \cite{Yelin2019} & 10.6\% & 33.6\%   \\
    \bottomrule
  \end{tabular}
\end{table}

\begin{table}[]
  \caption{Policy Comparison: Improvement in both outcomes}
  \label{tab:comparison-3}
  \centering
  \begin{tabular}{lcc}
    \toprule
     & \textbf{IAT} & \textbf{2nd-line usage}  \\
    \midrule
     Doctor & 11.9\% & 33.6\%\\
     Thresholding & 10.9\% & 0.8\%  \\
    Expected reward maximization & 10.8\% & 1.0\%   \\
    Direct learning & 10.7\% & 0.9\%   \\
    \bottomrule
  \end{tabular}
\end{table}

\begin{figure}
  \centering
  \includegraphics[width=\linewidth]{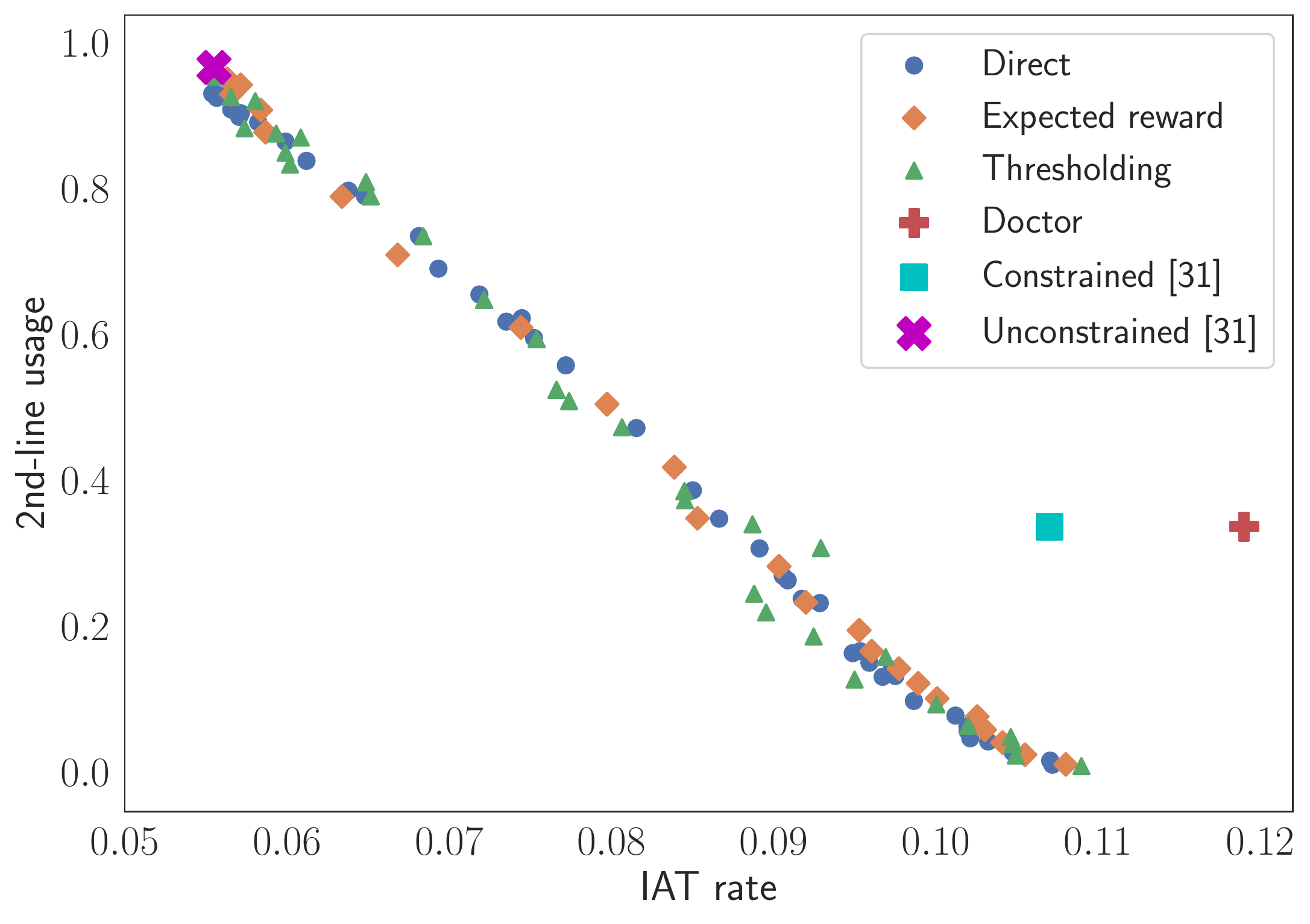}

  \caption{Performance frontier of learned policies compared to clinician performance and baselines. Policies with performance toward the bottom left of the figure are better.  Our three approaches - thresholding, expected reward maximization, and direct learning - outperform clinicians and the constrained baseline in \cite{Yelin2019}. Y-axis = 2nd-line usage rate, X-axis = inappropriate antibiotic therapy (IAT) rate.}
  \label{fig:frontier-main}
\end{figure}

We present the results of learning policies using both the indirect and direct policy learning approaches outlined in Section~\ref{sec:methods} for several settings of the cost constraints $b_j$ (for the thresholding approach) and reward weights $\omega$ (for the reward maximization and direct learning approaches). Figure~\ref{fig:frontier-main} shows the performance of the resulting set of policies for each approach on the 2014-16 patient cohort with respect to the IAT and 2nd-line usage rates.

We observe that the sets of policies learned by all three approaches achieve similar performance for a broad range of IAT and 2nd-line usage rates. In the reward maximization and direct learning approaches, the reward weight $\omega$ is able to successfully control the trade-off learned by the policy. As $\omega$ is reduced (i.e, treatment effectiveness is less important), the policy performance moves down and to the right along the frontier in Figure~\ref{fig:frontier-main}. 

We compare the performance of our learned policies to that of doctors on the same patient cohort in Tables \ref{tab:comparison-2} and \ref{tab:comparison-3}. In Table \ref{tab:comparison-2}, we choose a policy from the policy set constructed by each method that does no worse than clinicians on 2nd-line usage rate, and compare the corresponding IAT rates. All three approaches reduce the IAT rate by over 25\% relative to clinicians, while also producing a minor reduction in the 2nd-line usage rate. All approaches also outperform the constrained baseline from \cite{Yelin2019}, achieving a relative reduction in the IAT rate of over 15\%.
In Table \ref{tab:comparison-3}, we choose a policy from the policy set constructed by each method that improves both IAT and 2nd-line usage rates relative to clinicians. All three approaches are able to virtually eliminate 2nd-line usage while also reducing the IAT rate by nearly 10\% relative to clinicians. 

The unconstrained baseline proposed in \cite{Yelin2019} produces a policy achieving an extremely low IAT rate while using essentially only 2nd-line treatments. This is a product of the significantly lower resistance rates to 2nd-line treatments, which results in predicted resistance probabilities that are also almost always lower than predictions for 1st-line treatments. Such high usage of 2nd-line treatments would not be useful in practice.

Overall, we find that the frontier of policy performance for all three approaches lies significantly below and to the left of the point representing clinician performance. We can obtain policies that achieve significant improvements in both objectives of interest relative to clinical practice and previous baselines by selecting appropriate policies along the frontier. 

\subsection{Policy Interpretation}

\begin{table}[]
  \caption{Top features driving recommendation of NIT over SXT, both 1st-line agents. `Window' refers to the time window, in days, over which the feature was computed.  For instance, ``Prior resistance: SXT / 180'' is an indicator for resistance to SXT in the past 180 days.}
  \label{tab:features-1}
  \centering
  \begin{tabular}{lr}
    \toprule
     \textbf{Feature} & \textbf{Window} \\
    \midrule
    Prior resistance: SXT & 180 \\
    Prior resistance: SXT & All \\
    Prior resistance: SXT & 90 \\
    Prior treatment: SXT & 7  \\
    Prior treatment: Folate-inhibitor & 7  \\
    Location: Emergency Room & \\
    Prior treatment: Clarithromycin & 14 \\
     Prior resistance: Gentamicin & All \\
    Prior resistance: CIP  & 180  \\
    Prior treatment: SXT & 14  \\
    \bottomrule
  \end{tabular}
\end{table}

The direct policy learning approach enables interpretation of the learned treatment policy to understand features important for decision-making. The linear model learns a $m \times K $ weight matrix $\theta$, where each column contains the coefficients used to calculate the output for a particular antibiotic. 

We examine the features important in our policy's decisions for recommending antibiotic $a$ over antibiotic $a'$ by looking at pairwise differences in coefficient values in the corresponding columns of $\theta$. We extract the features $i$ for which $ \theta_{a}(i) - \theta_{a'}(i)$ is large. We can perform this comparison for all pairwise combinations of antibiotics in our action space to extract the features of interest.

\begin{figure*}[h]
  \centering
  \includegraphics[width=\textwidth]{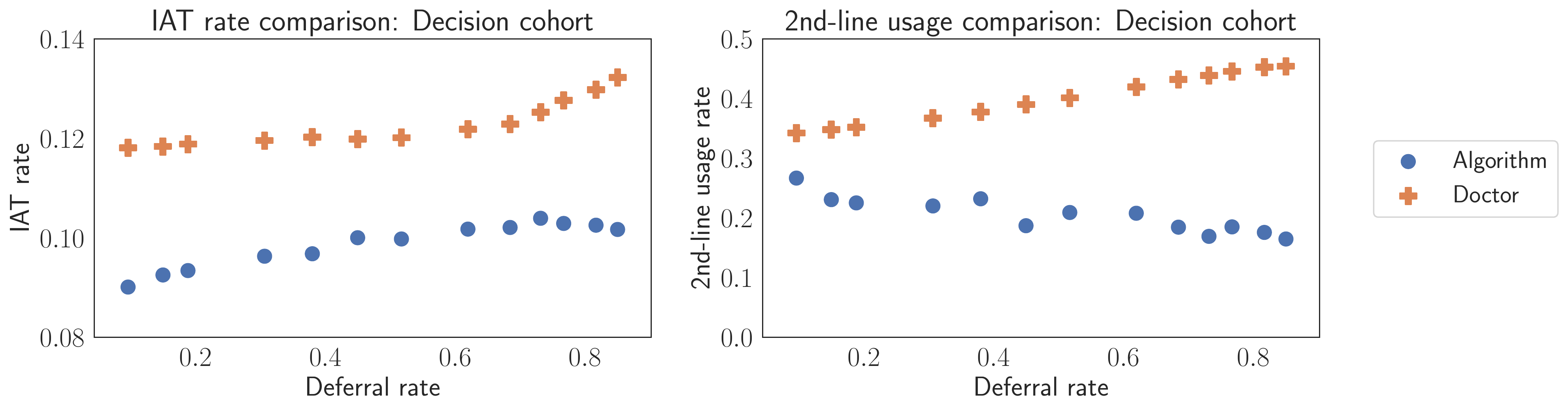}
  \caption{Comparison of doctor and algorithm IAT rates (left) and 2nd-line usage (right) within the `decision cohort' - the subset of examples where the model makes a decision - at different deferral rates. As the deferral rate increases above 40\%, the gap between clinician and algorithm performance in both objectives increases. The model learns to take action on the cases where clinicians exhibit the poorest performance.}
  \label{fig:decision-cohort-analysis}
\end{figure*}

In our analysis here, we specifically focus on the factors driving recommendation of NIT over SXT (both 1st-line antibiotics). The populations of patients that are resistant to NIT and SXT are largely disjoint (i.e, there are many patient resistant to exactly one of these agents), so accurately deciding when to use one agent over another is crucial for good policy performance. For a policy trained using a reward function with $\omega=0.88$, these features are listed in Table~\ref{tab:features-1}.

Many of these features align with knowledge that prior resistance or exposure to an antibiotic promotes future resistance to that treatment. For instance, prior resistance to SXT is an important factor in driving recommendation of NIT, as well as prior treatment with SXT in the recent past.  Almost all identified features are specific to an individual's history of antibiotic exposure and resistance, with the exception of one feature corresponding to the location at which the patient was seen. Population-level statistics related to antibiotic resistance and usage rates, such as colonization pressure or cumulative antibiotic usage, are not as useful for determining the selection of one 1st-line antibiotic over another. 

\subsection{Deferring to Doctors}
 In high-stakes settings such as healthcare, humans are a critical component of the decision-making process, and it is thus useful to learn models with the ability to defer to a human decision-maker. Several lines of recent work have developed methods for training models to defer in a way that optimizes the combined performance of humans and learned models \cite{madras2018predict, mozannar2020consistent, bansal2020optimizing, wilder2020learning}. 
 
 In medical settings such as antibiotic prescription, one might want the algorithm to only take decisions on cases where clinician decisions are particularly likely to result in ineffective treatment. Limiting the number of algorithmic interventions in this way may help doctors mitigate the well-documented phenomenon of `alert fatigue' resulting from an excessive number of computerized alerts \cite{nanji2018medication} and increase the likelihood that they incorporate algorithm input into their decision-making process.

The ability to incorporate deferral in a straightforward manner is a significant advantage of the direct learning approach. Adding this option in an indirect learning framework would require us to develop models of clinician behavior to calculate the expected reward of the doctor making the prescription decision for a given patient, which may be difficult to do. In this direct approach, incorporating a deferral option is no more difficult than incorporating an additional antibiotic treatment option. We simply expand our action space to include a `defer' action, and define the reward for this action as:
\begin{equation}
    r(\texttt{defer}) = r(a) + \lambda_{\texttt{defer}},
\end{equation}
where $a$ is the action taken by the doctor and $\lambda_{\texttt{defer}}$ is a positive parameter that controls the extent to which we incentivize deferral. Constructing the reward for deferral in this way encourages deferral on cases where the doctor takes a reward-maximizing action, and making a decision when doctors are likely to make an error. 

We examine the learned policy's performance for a fixed $\omega$ and several values of  $\lambda_{\texttt{defer}}$. In Figure~\ref{fig:decision-cohort-analysis}, we compare the IAT and 2nd-line usage performance of doctors and the learned policy on the `decision cohort', the subset of patients where the algorithm makes a decision. The results are shown for policies learned using a reward function with $\omega=0.92$ and values of $\lambda_{\texttt{defer}}$ in the range $[0.0, 0.10]$. 

We find that the gap in both IAT and 2nd-line usage rates between doctor and algorithm performance widens as the policy's deferral rate is increased and the algorithm takes decisions on fewer examples. As the deferral rate increases from 45\% to 85\%, the reduction in IAT rate on the decision cohort grows from 2.0\% to 3.1\%, a relative increase of over 50\%; the reduction in 2nd-line usage grows from 20\% to 29\%, a relative increase of 45\%. These trends indicate that the learned policy is successfully able to identify the subset of patients where it can provide the most improvement when constrained to only take action on a limited number of cases.

\subsection{Synthetic Evaluation: Direct vs. Indirect}

Even though direct and indirect methods achieved similar performance on our real-world antibiotic dataset, we demonstrate in this section a scenario where direct policy learning can significantly outperform an indirect approach. In particular, this can occur when the true treatment outcome models are complex, but the optimal treatment rule is simple.  For clarity, we illustrate the benefit of direct learning in a single-objective setting in these synthetic experiments, but an extension to multi-objective settings in the framework discussed previously is straightforward.

Our environment consists of feature vectors $\textbf{X} \in \mathbb{R}^{m}$ and an action space $\mathcal{A}$ with 3 actions. All feature values are drawn i.i.d. from a standard Gaussian distribution. We use the binary random variable $Y(a)$ to denote the outcome of action $a$. The values of each $Y(a)$ for a given $X$ are generated according to the following models:
\begin{equation}
    Y(a) \mid X  \sim \textbf{Bernoulli}\Big(\sigma\Big(X_a + \sum_{i=4}^{m}\alpha_iX_i^2  + \sum_{(i,j) \in S}\beta_iX_iX_j\Big)\Big)
\end{equation}
for $a=1,2,3$, where $\alpha_i, \beta_i$ are coefficients that are fixed across all 3 outcome models and $S$ is a subset of all distinct pairs of features. These are all nonlinear functions of the features $X$, but the Bayes-optimal treatment rule under these outcome models is given by an argmax over linear functions
\begin{equation}
    \pi^*(X) = \argmax_{a \in \lbrace 1, 2, 3 \rbrace } \ X_a
    \label{eq:bayes-optimal}.
\end{equation}
We compare the performance of an indirect approach (expected reward maximization) and the direct policy optimization approaches for policy learning in this environment. In the indirect approach, we independently train logistic regression models $h_a$ to predict the outcomes $Y(a)$ for each $a$. The treatment rule is then defined as $\argmax_{a} h_a(x)$. In the direct approach, we optimize the following loss function, where $f$ is parameterized by a linear model:
\begin{equation}
    \tilde{L}(f,x) = -\sum\limits_{i=1}^n \sum\limits_{a \in \mathcal{A}} \1{Y(a) = 1}\log \frac{\exp f_a(x)}{\sum_{a'} \exp f_{a'}(x)}.
\end{equation}

\begin{figure}
    \centering
    \includegraphics[width=\linewidth]{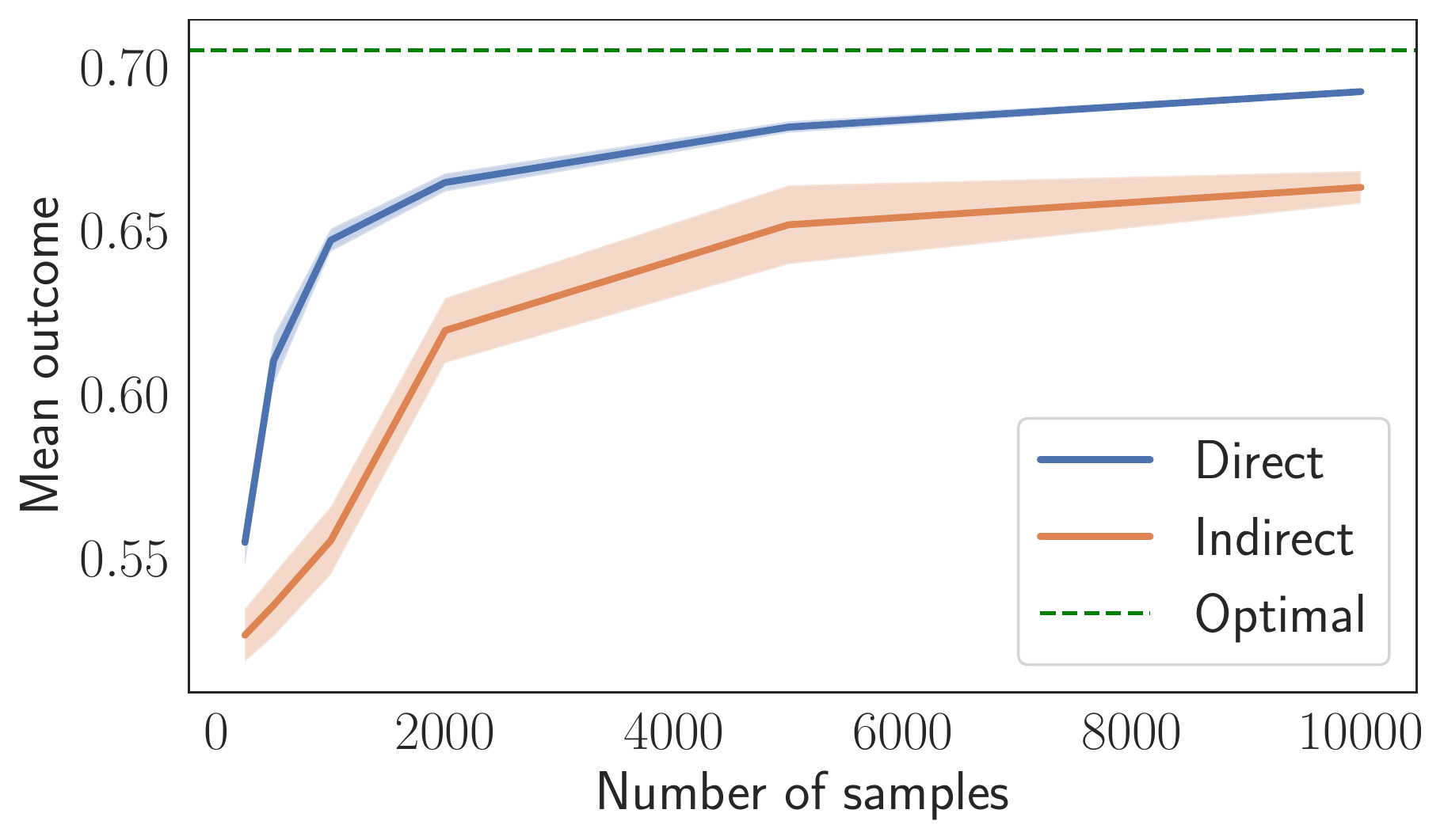}
    \caption{Comparison of indirect (expected reward maximization) and direct learning approaches in synthetic environment. The direct approach outperforms the indirect approach and approaches Bayes-optimal performance.}
    \label{fig:synthetic-exp-results}
\end{figure}

\noindent The results are shown in Figure~\ref{fig:synthetic-exp-results}. We plot the mean outcome of both approaches on a held-out test set for various training set sizes. We only compute the mean performance on the subset of examples in the test set for which outcomes were not uniform across all 3 actions (i.e, not all 0 or 1), as performance on the remaining samples does not depend on the policy. We also plot the mean performance of the Bayes-optimal policy given by Equation~\eqref{eq:bayes-optimal}. 

Direct policy learning significantly outperforms the indirect approach across a wide range of training set sizes and rapidly approaches the Bayes-optimal performance with far fewer samples. This synthetic experiment demonstrates that the direct learning approach, in contrast to the indirect approach, is able to take advantage of scenarios where the optimal treatment policy is simple, even when the true conditional outcome models are complex.

\section{Conclusion}

In this work, we presented three approaches for learning treatment policies in clinical settings with multiple treatment objectives and access to retrospective data that provides strong indicators for counterfactual treatment outcomes. We applied these approaches for learning antibiotic treatment policies for UTIs, and found that all our approaches achieved comparable performance and exceeded clinician performance across multiple treatment objectives. We also evaluated trade-offs among these approaches, and highlighted some significant advantages of direct learning in both real-world and synthetic experiments. We note that if one were in a setting without access to all counterfactual outcomes, one could still use the methods presented in this work in conjunction with appropriate estimators for the unobserved treatment outcomes.

Our empirical evaluation provides a real-world example for how to use machine learning to guide treatment selection. The fully observed setting has impactful applications in settings including antibiotic prescribing and precision medicine using patient-derived xenograft (PDX) models. At the same time, it is clearly a special case and is much simpler than the partially observed setting, where one has to simultaneously grapple with counterfactual estimation from biased data. As such, it is an ideal test bed for studying more subtle aspects of policy learning for guiding treatment selection. In this work, we addressed two important aspects of making treatment decisions in the real world: handling trade-offs between multiple objectives and deferring to clinician decisions when appropriate; future work using this type of data could examine the impact of nonstationarity on policy learning methods or techniques for developing more interpretable policies.

\section*{Acknowledgements}

This work was supported by a Massachusetts General Hospital -- Massachusetts Institute of Technology Grand Challenges Award (S.K., M.O., S.B., H.Z), a Harvard Catalyst CMeRIT grant (S.K.), and a National Science Foundation CAREER award \#1350965 (S.B., D.S.).  This work was also supported in part by Office of Naval Research Award No. N00014-17-1-2791 (M.O.).

\bibliographystyle{ACM-Reference-Format}
\bibliography{camera-ready.bib}
\include{appendix}

\end{document}

%% file: appendix.tex
\section*{APPENDIX}
\appendix
\section{Theoretical Results for Direct Policy Learning}

In this section we provide a self-contained proof of the consistency of our chosen loss function, known as the multinomial deviance loss \cite{Huang2019} in the literature on multi-category cost-sensitive classification with convex surrogates.    First, we note the following fact
\begin{thmprop}
  The function $\E_{r | x} \tL(f, x, r)$ is convex in $f$ for non-negative rewards $\mathbf{r}$.
  \label{prop:1}
\end{thmprop}
\begin{proof}
  The expectation $\E_{r | x}$ preserves convexity, so we just need to confirm that $\tL(f, x, r)$ is convex in $f$, which we can do so by rewriting as 
  \begin{equation}
    \tilde{L}(f, x, r) =  \sum^{}_{a \in \cA} r(a) \left[\left(\log \sum_{a'} \exp f_{a'}(x)  \right) - f_{a}(x) \right].
  \end{equation}
  The inner term is a convex function of $f$ because it is a non-negative sum of convex functions of $f$, namely log-sum-exp and $-f$.  The outer sum is a non-negative sum, since the rewards are specified to be non-negative, which preserves convexity.
\end{proof}
\begin{thmprop}
  For non-negative rewards $\mathbf{r}$, and for an $f^*$ that satisfies $f^* = \inf_f \E_{x, r} \tL(f, x, r)$, the corresponding policy $\pi^*(x) = \argmax_a f^*_a(x)$ is equivalent to the the Bayes-optimal policy $\pi^*(x) = \argmax_a \E[r(a) | X]$.
\end{thmprop}
\begin{proof}
  First, we can write this as 
  \begin{align*}
  \inf_f \E_{x, r} \tilde{L}(f, x, r) &= \E_{x} \inf_{f(x)} \E_{r|x} \tilde{L}(f, x, r).
  \end{align*}

  Because $\E_{r|x} \tL(f, x, r)$ is convex in $f$ (see Proposition~\ref{prop:1}), we just need to find a critical point where $\frac{\partial}{\partial f_{a^*}} \tL(f(x), x, r) = 0, \ \forall a^* \in \cA$.  We can see that 
  \begin{align*}
  &\frac{\partial}{\partial f_{a^*}} \E_{r|x} \tL(f, x, r) \\
  &= - \frac{\partial}{\partial f_{a^*}} \sum^{}_{a \in \cA} \E[r(a) | X] \log \frac{\exp f_a(x)}{\sum_{a'} \exp f_{a'}(x)} \\
  &= - \frac{\partial}{\partial f_{a^*}} \E[r(a^*) | X] \log \frac{\exp f_{a^*}(x)}{\sum_{a'} \exp f_{a'}(x)} \\
  &- \frac{\partial}{\partial f_{a^*}} \sum_{a \neq a^*} \E[r(a) | X] \log \frac{\exp f_{a}(x)}{\sum_{a'} \exp f_{a'}(x)}  \\
  &= - \E[r(a^*) | X] \left[ 1 - \frac{\exp f_{a^*}}{\sum \exp f_{a'}} \right] + \sum_{a \neq a^*} \E[r(a) | X] \frac{\exp f_{a^*}}{\sum_{a'} \exp f_{a'}}  \\
  &= -\E[r(a^*) | X]  + \frac{\exp f_{a^*}(x)}{\sum_{a'} \exp f_{a'}(x)} \sum_{a} \E[r(a) | X] = 0 \\
  \implies& \frac{\E[r(a^*) | X]}{\sum_a \E[r(a) | X]} = \frac{\exp f_{a^*}(x)}{\sum_{a'} \exp f_{a'}(x)}
  \end{align*}
  From this, we can see that 
  \begin{align*}
    \argmax_{a \in \cA} \E[r(a) | X] &= \argmax_{a \in \cA} \frac{\E[r(a) | X]}{\sum_{a'} \E[r_{a'} | X]} \\
     &= \argmax_{a \in \cA} \frac{\exp f_{a}}{\sum_{a'} \exp f_{a'}} \\
     &= \argmax_{a \in \cA} \log \left(\frac{\exp f_{a}}{\sum_{a'} \exp f_{a'}} \right) \\
     &= \pi^*(x)
  \end{align*}
  Completing the proof that at optimality, the optimal $$\fs = \arg\inf_{f(x)} \E_{r|x} \tL(f, x, r)$$ yields a calibrated decision rule $\pi^*(x)$.
  \end{proof}
  
We make two minor remarks:  First, as a practical matter, we drop the usual constraint (used to ensure uniqueness) that $\sum_a f_a(x) = 0$, as we impose $\ell_2$ regularization on the weights of our $f_a(x) = \theta^T_a x$ functions in our experiments.  Second, this formulation requires that the reward vector $\mathbf{r}$ is non-negative, but this can be relaxed in a straightforward way by replacing $r(a)$ with $\max_{a'} r(a') - r(a)$.  We tried this latter formulation in our experiments and it did not have a significant impact on results.

\section{Dataset}

\subsection{Feature Details}
In this section, we provide additional details about the construction of a few features used in the models.
\\\\ \textbf{Lab Values}. For a given lab test and backward window, the corresponding feature value is the mean result of all results for that lab value within the specified time window. The dataset contains lab results for white blood counts (WBC), absolute neutrophil counts (ANC), absolute lymphocyte counts (ALC), and CD4/CD8 counts.
\\\\ \textbf{Colonization pressure}. Colonization pressure is defined as the proportion of resistant specimens to a given antibiotic across a specified location and time window. We calculate the colonization pressure for 25 different antibiotics in the window from 7 days prior to 90 days prior to the specimen collection date among all patients with UTIs. We calculate colonization pressure values at 3 different location hierarchies: (1) the ward/clinic of specimen collection, (2) the hospital of collection, and (3) across the entire dataset.

\section{Experiments}
This section contains details about the experiments conducted in this paper. Further implementation details can be found in the Github repository for this paper at \url{https://github.com/clinicalml/fully-observed-policy-learning}.

\subsection{Thresholding}
We use \texttt{sklearn}'s logistic regression implementation to learn models for predicting the effectiveness of each of the four antibiotic treatments. Hyperparameters are tuned by constructing twenty 70\%/30\% train/validation splits of the full training set, and selecting the parameters that maximize the average AUC on the 20 validation splits. The test AUCs achieved for each of the antibiotics with these optimal hyperparameters are shown in Table~\ref{tab:test-aucs}. We also explored using nonlinear predictive models (e.g., random forests, XGBoost), but found that these model classes did not significantly improve predictive performance or policy performance on the validation set and chose not to use them for our final analyses.

\begin{table}
  \caption{Test AUCs for predicting treatment effectiveness}
  \label{tab:test-aucs}
  \centering
  \begin{tabular}{l|cccc}
    \toprule
     \textbf{Antibiotic} & NIT & SXT & CIP & LVX  \\
     \textbf{AUC} & 0.563 & 0.593 & 0.637 & 0.637 \\
    \bottomrule
  \end{tabular}
\end{table}

Our threshold search space is defined implicitly by a fixed set of 11 false negative rates as follows: for each FNR value and antibiotic, the corresponding probability threshold is the one that achieves that FNR rate among the training set resistance predictions for that drug. Since there is a strong correlation between resistance to CIP and LVX, we constrain $\cT$ to combinations where thresholds for CIP and LVX are the same. Our threshold space $\cT$ consists of $11^3 = 1,331$ possible combinations.

When a threshold combination results in predictions of resistance for all antibiotic treatments, the policy falls back on a default 1st-line antibiotic. We chose to always default to recommending NIT, as it has a significantly lower overall resistance rate than SXT in the training set. The optimal threshold combination for a given value of the budget constraint $b_j$ is selected by computing the average policy performance across the same twenty validation sets used for tuning hyperparameters of predictive models. In Figure 1, we show the results of optimal policies $\pi_j$ for budget constraint values $b_j$ in [0.01, 0.05] (in increments of 0.01) and (0.05, 1.0] (in increments of 0.025). The performance of each policy is computed as the mean IAT and 2nd line usage rates of 20 samples bootstrapped with replacement from the test set, where each sample is the same size as the full test set.

\subsection{Expected Reward Maximization}
The procedure for training and tuning logistic regression models is the same as described for the thresholding approach. The indirect learning policy frontier in Figure 1 contains the performance of models learned using values of $\omega$ in the range [0.85, 1], computed as the mean IAT and 2nd line usage rates of 20 samples bootstrapped with replacement from the test set.

\subsection{Direct Learning}
Models are trained for 50 epochs using an Adam optimizer with a learning rate of $10^{-4}$ and L2 regularization with penalty $3 \times 10^{-3}$.  The policy frontier in Figure 1 contains the performance of models learned using values of $\omega$ in the range [0.85, 1]. We plot the mean IAT and 2nd line usage outcomes for each setting of $\omega$ from policy learning across 20 trials.
 
The direct learning model with the deferral action was trained using the same learning rate and regularization hyperparameters. The values for $\lambda_{\texttt{defer}}$ were selected from the range $[0,0.10]$. Mean deferral rates and IAT / 2nd line usage rates shown in Figure~\ref{fig:decision-cohort-analysis} were calculated over 100 trials for each setting of $\lambda_{\texttt{defer}}$.

\subsection{Baselines}

We implement the constrained and unconstrained treatment selection approaches presented in \cite{Yelin2019}. In both approaches, we first learn conditional outcome models for treatment resistance $f_a(x)$ for each treatment $a$ using the same training procedure as described for our indirect approaches (Section C.1). The unconstrained treatment selection approach selects the treatment with the lowest predicted resistance probability (i.e., highest predicted treatment effectiveness) across the trained outcome models.

The constrained approach adjusts the predicted resistance probabilities by adding treatment-specific costs $c_a$:
\begin{align*}
    f'_a(x) = f_a(x) + c_a.
\end{align*}

The policy $\pi_\text{C}$ is defined by selecting the treatment with the minimal adjusted `score':
\begin{align*}
     \pi_{\text{C}}(x) &= \argmin_{a}f'_a(x).
\end{align*}

The costs $c_a$ are chosen to constrain the treatment distribution of $\pi_{\text{C}}$ to match the empirical treatment distribution of clinicians. We solve for the costs by iteratively updating them according to the following equation until convergence:
\begin{align*}
    c^{t+1}_a \leftarrow c^t_a + \alpha \cdot (\text{Count}^{\pi^t}(a) - \text{Count}^{\text{doc}}(a)),
\end{align*}
where $\alpha$ is a step size, $\text{Count}^{\pi^t}(a)$ is the number of uses of treatment $a$ in the policy defined by the costs at step $t$, and $\text{Count}^{\text{doc}}(a)$ is the number of clinician uses of treatment $a$. 

\subsection{Synthetic Experiments}
The synthetic environment consists of a 10-dimensional feature space and an action space $\cA$ with 3 actions. Each feature value is drawn i.i.d. from a standard normal distribution. The feature coefficient values (i.e, $\alpha_i, \beta_i$) are selected manually to ensure that the mean outcomes for each action in the dataset are roughly 0.5. All coefficients have magnitude larger than 1 to ensure that learning these values is necessary for learning a good predictive model.

In the indirect approach, we train logistic regression models to predict treatment outcomes for each action using the \texttt{saga} solver in \texttt{sklearn}'s implementation. Hyperparameters - L1 vs. L2 regularization and regularization strength - are tuned using 10-fold cross validation on the training set. Models are trained for a maximum of 100 iterations. 

In the direct approach, the convex surrogate loss is optimized using SGD with a learning rate of 0.1 and L2 regularization with $\lambda=0.001$. Models are trained for 50 epochs.

Figure~\ref{fig:synthetic-exp-results} shows the outcomes of indirect and direct policy learning using training sets of various sizes on a fixed test set of $10^6$ samples drawn from the specified generative model. We only evaluate outcomes on samples where there was at least one ineffective and one effective treatment (i.e, not all 0 or 1 outcomes), as these are the only examples where the policy's decision can affect the outcome. We train both indirect and direct approaches on the same 25 randomly drawn training sets for each sample size, and plot the mean outcome and standard deviations for each setting across these samples in Figure~\ref{fig:synthetic-exp-results}.